\documentclass{sig-alternate-05-2015}

\begin{document}





%

\title{Nrityantar: Pose oblivious Indian classical dance sequence classification system}

%
%
%
%
%

\numberofauthors{3} 
%
\author{
%
%
\alignauthor
Vinay Kaushik\thanks{Equal contribution}\\
       \affaddr{Dept. of Electrical Engineering}\\
       \affaddr{IIT Delhi}\\
       \email{eez158117@ee.iitd.ac.in}
\alignauthor
Prerana Mukherjee\footnotemark[1]\\
       \affaddr{Dept. of Electrical Engineering}\\
       \affaddr{IIT Delhi}\\
       \email{eez138300@ee.iitd.ac.in}
\alignauthor
Brejesh Lall\\
       \affaddr{Dept. of Electrical Engineering}\\
       \affaddr{IIT Delhi}\\
       \email{brejesh@ee.iitd.ac.in}\\
}

\maketitle
\begin{abstract}
In this paper, we attempt to advance the research work done in human action recognition to a rather specialized application namely Indian Classical Dance (ICD) classification. The variation in such dance forms in terms of hand and body postures, facial expressions or emotions and head orientation makes pose estimation an extremely challenging task. To circumvent this problem, we construct a pose-oblivious shape signature which is fed to a sequence learning framework. The pose signature representation is done in two-fold process. First, we represent person-pose in first frame of a dance video using symmetric Spatial Transformer Networks (STN) to extract good person object proposals and CNN-based parallel single person pose estimator (SPPE). Next, the pose basis are converted to pose flows by assigning a similarity score between successive poses followed by non-maximal suppression. Instead of feeding a simple chain of joints in the sequence learner which generally hinders the network performance we constitute a feature vector of the normalized distance vectors, flow, angles between anchor joints which captures the adjacency configuration in the skeletal pattern. Thus, the kinematic relationship amongst the body joints across the frames using pose estimation helps in better establishing the spatio-temporal dependencies. We present an exhaustive empirical evaluation of state-of-the-art deep network based methods for dance classification on ICD dataset.
\end{abstract}

%
%


%
%

%
%



\section{Introduction}
\label{sec:intro}
Research in action recognition from video sequences has expedited in the recent years with the advent of large-scale data sources like ActivityNet\cite{caba2015activitynet} and ImageNet \cite{krizhevsky2012imagenet} and availability of high computing resources. Along with other computer vision areas, Convolutional neural networks (ConvNets) has been extended for this task as well by utilizing the spatio-temporal filtering approaches \cite{ji20133d}, multi-channel input streams \cite{zolfaghari2017chained,karpathy2014large}, dense trajectory estimation with optical flow \cite{cheron2015p}. However, the extent of success is not that overwhelming as is in the case of image classification and recognition tasks. Long Short Term Memory networks (LSTMs) capture the temporal context such as pose and motion information effectively in action recognition \cite{wang2014action}. We attempt to advance the research work done in human action recognition to a rather specialized domain of Indian Classical Dance (ICD) classification.

There is a rich cultural heritage prevalent in the Indian Classical Dance or Shashtriya Nritya\footnote{``Shastriya Nritya" is the Sanskrit equivalent of ``Classical dance".}  forms. The seven Indian classical dance forms include: Bharatanatyam, Kathak, Odissi, Manipuri, Kuchipuri, Kathakali and Mohiniattam. Each category varies in the hand (or hasta-mudras) and body postures (or anga bhavas), facial expressions or emotions depicting the nava-rasas\footnote{``Rasa" is an emotion experienced by the audience created by the facial expression or the feeling of the actor. ``Nava" refers to nine of such emotions namely, erotic, humor, pathetic, terrible, heroic, fearful, odious, wondrous and peaceful.}, head orientation, as well as the rhythmic musical patterns accompanied in these. Even the dressing attires and make-up hugely differs across them. All these put together constitute the complex rule engines which govern the gesture and movement patterns in these dance forms. In this work, however we do not address the gesture aspect of these complex dance forms. In this paper, we classify these categories based on the human body-pose estimation utilizing sequential learning.  Instead of feeding a simple chain of joints in the sequence learner which generally hinders the network performance, we constitute a feature vector of the normalized distance vectors which captures the adjacency configuration in the skeletal pattern. Thus, the kinematic relationship amongst the body joints across the frames helps in better establishing the spatio-temporal dependencies.

\begin{figure}[h]
\centering
\fbox{
\includegraphics[scale=0.22]{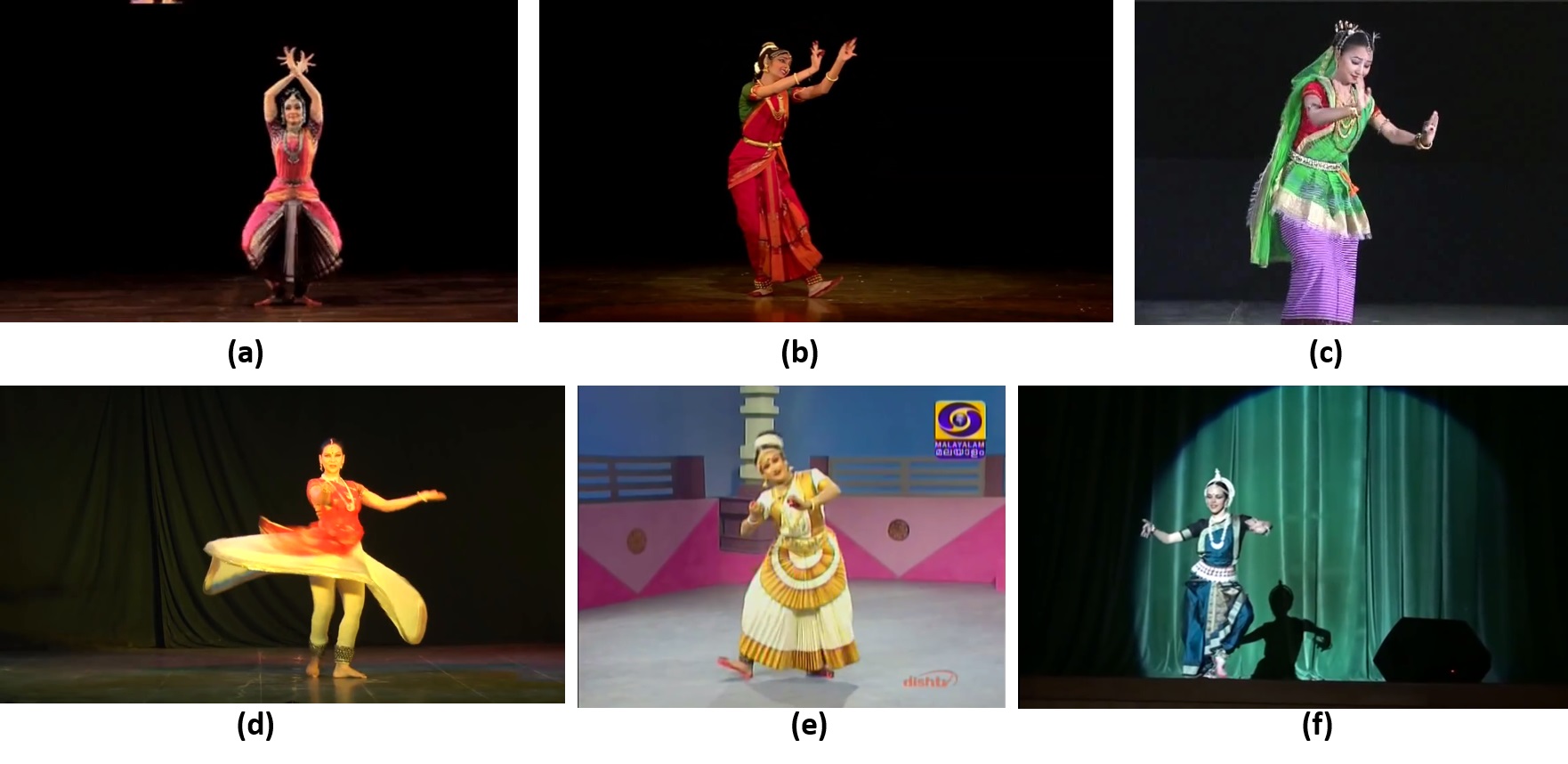}}
\caption{Indian Classical Dance (ICD) Forms: 
(a) Bharatnatyam (b) Kuchipudi (c) Manipuri (d) Kathak 
(e) Mohiniattam   (f) Odissi}
\label{fig:danceform}
\end{figure}
During an Indian classical dance performance, the dancer may be often required to sit in a squat position or half-sitting pose, cross-legged pose, take circular movements or turns, thus leading to severe body-occlusion which renders the human pose estimation as a highly challenging task. The addition of certain dress attires further complicates the pose estimation. As in the case of Manipuri dance, women dancers wear an elaborately decorated barrel shaped long skirt which is stiffened at the bottom and closed near the top due to which the leg positions are obscured. In some dance forms such as Kathakali, only the facial expressions are highlighted and the dancers wear heavy masks. In this work, we leverage the feature strength by constructing dance pose shape signature which is fed to a sequence learning framework. It encodes the motion information along with the geometric constraints to handle the aforementioned problems. In this paper, we address six oldest Indian dance classes namely Bharatnatyam, Kathak, Odissi, Kuchipudi, Manipuri and Mohiniattam. Bharatnatyam is one of most popular ICD belonging to the southernbelt of India, particularly originated in Tamil Nadu. Kuchipudi emanated from Andhra Pradesh and Telangana regions. Mohiniattam belongs to Kerala. Kathak originated in the northern part of India. Manipuri and Odissi are from eastern part of India, Manipur and Orissa respectively. We have not considered Kathakali ICD as it involves mainly facial expressions and does not contain enough visual dance stances to be catered in the adopted classification pipeline. The different dance forms have been depicted in Fig. \ref{fig:danceform}.

Remaining sections in the paper are organized as follows. In Sec. \ref{sec:related} we discuss the related work in dance and action recognition classification. In Sec. \ref{sec:proposed}, we outline the methodology we propose to detect the person dancing, track the dancer's pose and movement as well as characterize the dancer's trajectory to understand their behavior. In Sec. \ref{sec:results}, we discuss experimental results and conclude the paper in Sec. \ref{sec:conclusion}.

\section{Related Works}
\label{sec:related}
\subsection{Handcrafted feature based approaches}
Most of the traditional works in action recognition domain constitute of constructing handcrafted features to discriminate action labels \cite{lecun1998gradient, wang2013action, laptev2005space, bregonzio2009recognising, dalal2005histograms}. Next, these features were encoded into Bag of Visual Words (BoVW) or Fisher Vector (FV) encodings. The classifier models are then trained on these encodings and classify the videos with suitable labels. However, these feature sets were not optimized and failed to capture the temporal dependencies in the video streams to encode the high-level contextual information. Histogram of Oriented Gradients \cite{dalal2005histograms} have remained the de-facto choice in capturing the shape information in action recognition task. Histogram of Optical Flow magnitude and orientation (HOF) \cite{dalal2006human} models the motion. They obtain better performance since they represent the dynamic content of the cuboid volume. Space time interest point (STIP) detectors \cite{laptev2005space} are extensions of 2D interest point detectors that incorporate temporal information. Similarly, as in the 2D case, STIP features are stable under rotation, viewpoint, scale and illumination changes. Spatio-temporal corners are located in region that exhibits a high variation of image intensity in all three directions(x,y,t). This requires that such corner points are located at spatial corners such that they invert motion in two consecutive frames (high temporal gradient variation). They are identified from local maxima of a cornerness function computed for all pixels across spatial and temporal scales. Holistic representation of each action sequence is done by a vector of features. Motion history images \cite{bobick2001recognition} obtained utilizing Hu moments are utilized as global space-time shape descriptors.

\subsection{Deep feature based approaches}
With the deep learning networks, most of the computer vision tasks have reached remarkable accuracy gains. Human action recognition can be solved more effectively if the entire 3D pose comprehensive information is being exploited in the deep networks \cite{ji20133d, wang2016temporal}. In \cite{zhu2016co}, authors utilize the co-occurrence of the skeletal joints in a sequence learning framework. However, since there is a joint optimization framework involved it is computationally expensive. In \cite{simonyan2014two}, authors investigate a two-stream ConvNet architecture and jointly model the spatio-temporal information using separate networks to capture these. They also demonstrate that multi-frame dense optical flow results in improved performance gains in spite of limited training data. In \cite{wang2015action}, authors utilize trajectory constrained pooling to aggregate the discriminative convolutional features into trajectory-pooled deep-convolutional descriptors. They construct two normalization methods for spatiotemporal and channel information. Trajectory-pooled deep-convolutional descriptor performs sampling and pooling on aggregated deep features thus improving the discriminative ability. 

\subsection{Pose feature based approaches}
Neural Network (CNN) architectures \cite{bulat2016human, newell2016stacked} can provide excellent result for 2D human pose estimation with images. However, handling occlusion is quite challenging task. In \cite{rogez2017lcr}, authors obtain pose proposals by localizing the pose classes using anchor poses. Then, the pose proposals are scored to get the classification score and regressed independently. Since, pose regression techniques result in a single pose estimation thus cannot handle multimodal pose distributions. In order to capture such information, the pose estimates can be discretized into various bins followed by pose classification. In \cite{mahendran2018mixed}, authors construct a classification-regression framework that utilizes classification network to generate a discretized multimodal pose estimate followed by regression network to refine the discretized estimate to a continuous one. The architecture incorporates various loss functions to enable different models. In \cite{zolfaghari2017chained}, authors incorporate pose, motion and color channel information into a Markovian model to incorporate the spatial and temporal information for localization and classification.

\section{Proposed Methodology}
\label{sec:proposed}
In this section, the proposed framework and its main components are discussed in detail including the recognition of a dance move from a sequence of frames in a video using LSTM networks and feature extraction for the video frames. First, we extract features using uniform frame skipping in a sequence of frames such that the skipping of frame does not affect the sequence of the action in the video. The feature vector comprises of multiple components. We perform feature extraction using a 3-tier framework combining Inception V3 features, 3D CNN features and novel pose signatures. The second component trains our model for dance classification. It takes a chunk of features as input where a chunk is defined by a collection of features for the selected frames in the video. The feature chunks are then fed to a LSTM network in order to classify various dance forms. The output of the LSTM layer is connected to 2 fully connected layers with Batch Normalization at each stage and is finally connected to a softmax layer of size 6, corresponding to the 6 ICD dance forms.

\begin{figure*}[h]
\centering
\fbox{
\includegraphics[scale=0.6]{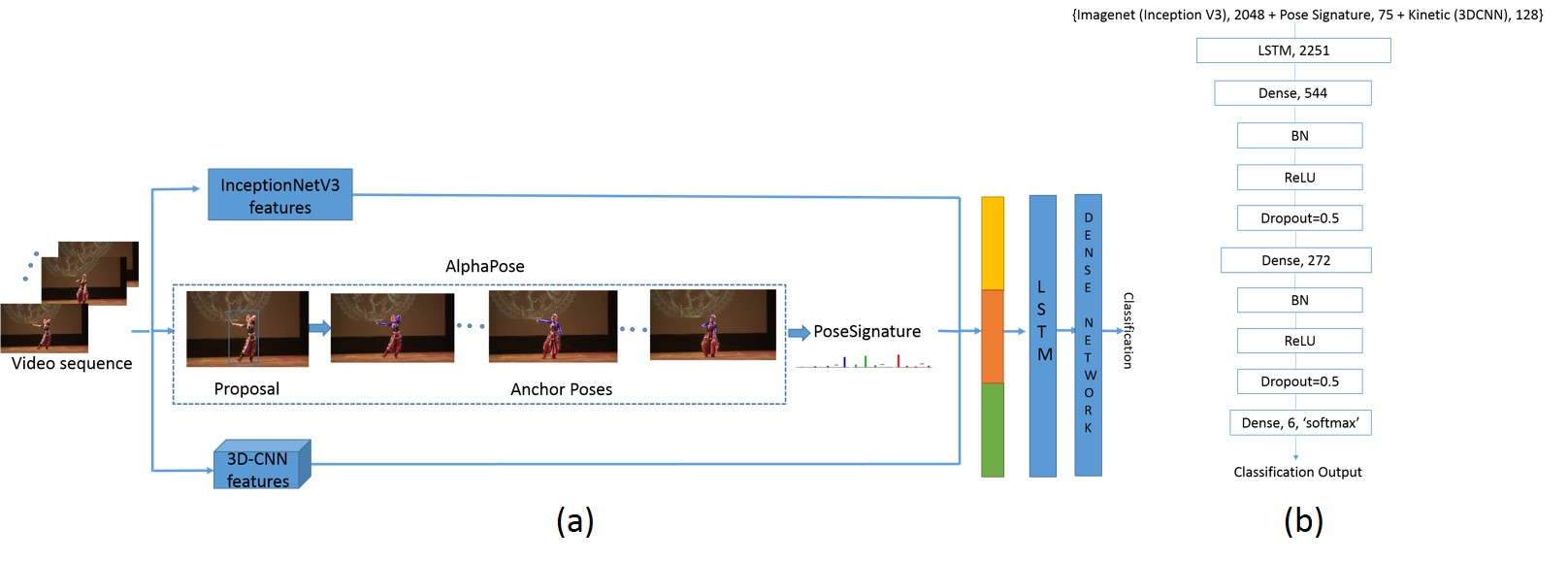}}
\caption{(a) Achitecture of the Nrityantar framework (b) Block Diagram of the sequence learning framework in Nrityantar. The number in the layers indicate the number of features.}
\label{fig:archi}
\end{figure*}
\subsection{Feature Construction: Inception V3 Features, 3D CNN, Pose Signature}
The feature vector for training our LSTM model is an embedding of various deep features (both 2D and 3D) and geometric cues in a single vector learning to classify the classical dance type. The 2D spatial cues are learnt using InceptionV3 network weights trained on ImageNet dataset \cite{deng2009imagenet}. The 3D temporal cues are described by using ResNext-101 with cardinality 32 on Kinetics dataset \cite{kay2017kinetics} which contains 400 video classes. In order to construct the pose signature, geometric information is used to construct features comprising of pose, flow and other structural information. The variation in Indian classical dance forms in terms of hand and body postures, head orientation makes pose estimation an extremely challenging task. To circumvent this problem, we construct a pose-oblivious shape signature which is fed to a sequence learning framework. The pose estimation is done as a two-fold process. The pose estimation is done using Alphapose framework. First, we represent person-pose in a frame of a dance video using a symmetric Spatial Transformer Networks (STN) to extract good person object proposals and CNN-based parallel single person pose estimator (SPPE). Next, the pose basis are converted to pose flows by assigning a similarity score between successive poses followed by non-maximal suppression. The video frames are fed into Alphapose framework computes the skeletal pose comprising of anchor joints at various body parts of a human being. The anchor joints are used to further compute various geometric features, resulting into a pose signature (as explained in Sec. \ref{sec:pose}. The feature generation strategy is described in detail further below.

Inception V3 features: These features widely are used for classification of images. The input size is $299$x$299$x$3$ and the output layer gives a feature of size $1$x$1$x$2048$. We utilize the pre-trained ImageNet model used for image classification. This has several advantages over the other networks. It incorporates Batch Normalization to the previous architectures, replaces the $5$x$5$ convolution layers in Inception V2 by two $3$x$3$ convolution layers and has the benefit of an added BN-Auxiliary layer i.e. the fully connected layer of the auxiliary classifier is also normalized, not just the convolutional layers.

ResNext-$101$ Kinetics features: It is pre-trained on Kinetic dataset with $400$ video classes. We have used $16$ consecutive frames for generating a $3D$ feature. The architecture comprises of $101$ convolutional layers with skip connections and cardinality of $32$. The output can be represented using a feature vector of size $2048$ or $16$x$128$ for $16$ frames. Thus, it gives a $128$-dimensional feature vector per frame.

\subsection{Pose Signature}
\label{sec:pose}
\begin{figure}[h]
\centering
\fbox{
\includegraphics[scale=0.28]{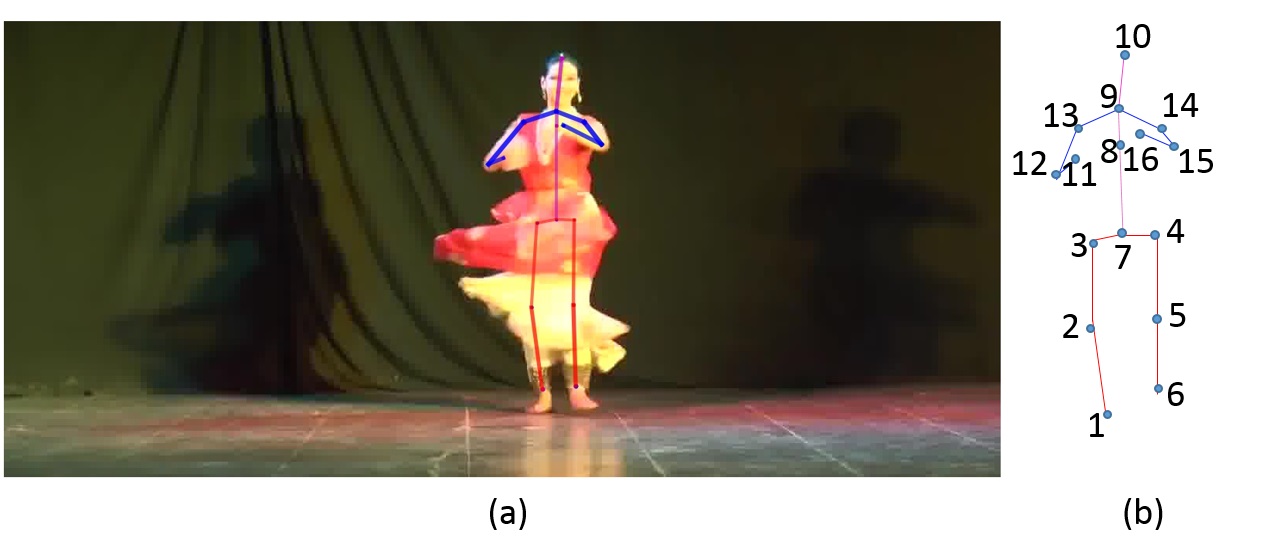}}
\caption{(a) Pose of a Kathak dancer (b) Visualization of the anchor joints}
\label{fig:pose}
\end{figure}

After the AlphaPose estimation, we construct novel pose signature which consists of the normalized distances and angles between the anchor joints in the pose vector. The pose vector constitutes 16 anchor joints namely, $1$: $foot_{right}$, $2$: $knee_{right}$, $3$:$hip_{right}$,
$4$:$hip_{left}$  ,    $5$: $knee_{left}$  ,    $6$: $foot_{left}$  ,    $7$: $hip_{center}$  ,    $8$:$spine$, 
$9$: $shoulder_{center}$,   $10$: $head$ ,   $11$: $hand_{right}$  ,$12$: $elbow_{right}$,$13$: $shoulder_{right}$  ,$14$: $shoulder_{left}$, $15$: $elbow_{left}$, $16$: $hand_{left}$ as shown in Fig. \ref{fig:pose}. We utilize joint $7$ as the reference point for taking the normalized distances to all other anchor joints. The distance metric used is Euclidean distance norm. We also incorporate the angles between the key anchor joints to characterize the dance stances. The angles are considered between these ordered anchor joint pairs: $\{1,3\}$,$\{2,7\}$,$\{4,6\}$,$\{5,7\}$,$\{11,13\}$, $\{9,12\}$,$\{9,15\}$,$\{14,16\}$. We embed the normalized distances between the leg joints ($\{1,6\}$,$\{2,5\}$) and hand joints ($\{11,16\}$, \\$\{12,15\}$)  as well in the pose signature. Next, we calculate the flow vectors between the anchor joints in successive frames to characterize the temporal dependencies. We also compute the flow directionality for the anchor joints across successive frames. Thus, in total the pose signature constitutes a $75D$ vector. Instead of simply feeding a chain of joints in the sequence learner which generally hinders the network performance we constitute a pose signature which captures the adjacency configuration in the skeletal pattern. Hence, the kinematic relationship amongst the body joints across the frames using pose estimation step helps in better establishing the spatio-temporal dependencies.

\subsection{Sequence Learning Framework}
The feature vectors obtained from 3DCNN, pose signature and ImageNet features are concatenated to be fed into a sequence learning framework. We utilize Long Short Term Memory networks for dance sequence classification. Usually, Recurrent Neural Networks (RNNs) are difficult to train with various activation functions such as tanh and sigmoid due to the problems of vanishing gradient and error exploding \cite{hochreiter2001gradient}. LSTMs can be used to learn the long-term dependencies in place of RNNs and it solves the aforementioned problems. It consists of one self-connected memory cell $c$ and three multiplicative units, input gate $i$, forget gate $f$ and output gate $o$. 

Given an input sequence $x=(x_0,...,x_T )$, the activations of the memory cell and three gates are given as follows:
\begin{align}
    & i_t=\sigma(W_{xi} x_t+W_{hi} h_{t-1}+W_{ci} c_{t-1}+b_i)\\
   & f_t=\sigma(W_{xf} x_t+W_{hf} h_{t-1}+W_{cf} c_{t-1}+b_f )\\
   & c_t=f_t c_{t-1}+i_t (\tanh(W_{xc}  x_t+W_{hc} h_{t-1}+b_c)\\
   & o_t=\sigma(W_{xo} x_t+W_{ho} h_{t-1}+W_{co} c_t+b_o)\\
   & h_t=o_t  \tanh(c_t),
\end{align}

 where $\sigma(.)$ is the sigmoid function, W are the connection weights between two units and h denotes the output values of a LSTM cell.

We have fed input feature vector into the LSTM network. The concatenated feature is $2251D$ vector. We have utilized a dense network after an LSTM layer followed by softmax layer for dance sequence classification as shown in Fig. \ref{fig:archi}(b). The constructed feature vector is able to characterize the spatio-temporal dependencies between the anchor joints. We utilize the cross-entropy loss function given as below.

\textbf{Training Loss:}
We use Adam optimizer with Categorical cross entropy for training. The equation for categorical cross entropy is:
\begin{equation}
    \frac{-1}{N}\sum_{i=1}^N \sum_{c=1}^C 1_{y_i\epsilon C_c} log  p_{model} [{y_i}\epsilon{C_c}]
\end{equation}

The double sum is over the observations (video features) $i$, whose number is $N$, and the categories $c$, whose number is $C$. The term $1_{y_i\epsilon C_c}$  is the indicator function of the $i^{th}$ observation belonging to the $c^{th}$ category. The $p_{model}$ $[{y_i}\epsilon{C_c}]$  is the probability predicted by the model for the $i^{th}$ observation to belong to the $c^{th}$ category. When there are more than two categories, the neural network outputs a vector of $C$ probabilities, each giving the probability that the network input should be classified as belonging to the respective category.

\section{Experimental Results}
\label{sec:results}
In this section, we evaluate our model and compare with other five different architectures on a benchmark dataset for Indian classical dance: ICD dataset \cite{samanta2012indian}. We also discuss the overfitting issues and the computational efficiency of the proposed model. 

\subsection{Evaluation Dataset and Parameter Settings}
ICD dataset mainly consists of curated videos from YouTube of primarily $3$ oldest dance forms Bharatnatyam, Kathak and Odissi with video class annotation. Each class consists of $30$ video clips of maximum resolution $400$×$350$ and of $25$ sec maximum duration. We have done data augmentation with these videos for classes Manipuri, Kuchipudi and Mohiniattam from YouTube. During data processing, we further clipped the video segments into $5$-$6$ seconds chunks of frames at $25$ fps to generate a maximum of $150$ frames. The train to test ratio for evaluation has been selected as $7:3$. The resultant dataset posed several challenges including varying illumination changes, shadow effects of dancers on stage, similar dance stances etc. The low accuracy of the skeleton joint coordinates and the partial body parts missing in some sequences makes this dataset very challenging. Tab. \ref{tab:3} shows the parameter settings of our proposed model on the evaluated dataset. For training, we utilize Adam optimizer with $0.0004$ as learning rate and $0.000001$ as decay. The loss function is categorical cross-entropy. Training is done for a maximum of $100$ epochs with early stopping if validation error is not improved for consecutive $5$ epochs. 

\begin{table}[]
\label{tab:3}
\caption{Parameters used for training the LSTM}
\begin{tabular}{|c|c|}
\hline
\multicolumn{2}{|c|}{\textbf{\begin{tabular}[c]{@{}c@{}}TRAINING\\ PARAMETERS\end{tabular}}}                                                                            \\ \hline
\textbf{Optimizer}                                                                       & Adam                                                                         \\ \hline
\textbf{Loss}                                                                            & Categorical Cross-entropy                                                    \\ \hline
\textbf{Learning rate}                                                                   & 0.0001                                                                       \\ \hline
\textbf{Decay}                                                                           & 0.000001                                                                     \\ \hline
\textbf{Feature length}                                                                  & 2251                                                                         \\ \hline
\textbf{Output length}                                                                   & 6                                                                            \\ \hline
\textbf{\begin{tabular}[c]{@{}c@{}}Sequence Length\\ (One training sample)\end{tabular}} & \begin{tabular}[c]{@{}c@{}}48 frames \\ (Uniformly Distributed)\end{tabular} \\ \hline
\textbf{Batch Size}                                                                      & 32                                                                           \\ \hline
\textbf{Maximum Epoch}                                                                   & 100                                                                          \\ \hline

\end{tabular}
\end{table}

\subsection{Evaluation Results and Discussion}
In order to verify the effectiveness of the proposed network, we compare with other four deep learning-based architectures: Convolutional LSTM (LRCN), CONV3D, Multilayer Perceptron (MLP) and LSTM. We also provide a baseline with the various features from Inception V3 ImageNet features, Kinetics 3D-CNN features and a combination of these with Pose signature using LSTM framework. The LSTM output is forwarded to 2 fully connected layers in all cases. The size of the layer varied as the feature length varied. The first dense layer is of quarter size to the LSTM output. The second dense layer is $1/2$ of the first one. This is done for both optimality as well as for a fair comparison between all features. The final layer is a dense softmax layer with size $6$.
Tab. \ref{tab:1} provides the classification results of the proposed architecture. Tab. \ref{tab:2} shows the confusion matrix for the 6 ICD forms. For training, we utilize Inception V3 pretrained model on ImageNet and extract image features. We randomly sample 48 frames from every video and pass those frames through ImageNet pretrained model to create a stacked feature set for each video. The initial input size to the model is (batch size, sample size, feature size) i.e. ($32$,$48$,$2048$). We add $2$ fully connected layers to the output of LSTM and feed it to softmax layer for evaluation. We achieved best results with LSTM framework as compared to other evaluated deep learning architectures. Second best results are obtained using Multi-layer Perceptron approach with reasonable accuracy ($40$-$50\%$) but lagging behind LSTM ($60$-$70\%$).

\begin{table}[]
\tiny	
\label{tab:1}
\caption{Final results}
\begin{tabular}{|c|c|c|c|c|c|}
\hline
\textbf{\begin{tabular}[c]{@{}c@{}}Dance \\ Class\end{tabular}} & \textbf{Precision} & \textbf{Recall} & \textbf{F1-score} & \textbf{Support} & \textbf{\begin{tabular}[c]{@{}c@{}}Class\\ Accuracy\end{tabular}} \\ \hline
\textbf{Bharatnatyam}                                           & 0.54               & 0.87            & 0.66              & 76               & 86.84                                                             \\ \hline
\textbf{Kathak}                                                 & 0.65               & 0.88            & 0.74              & 58               & 87.93                                                             \\ \hline
\textbf{Kuchipudi}                                              & 0.82               & 0.71            & 0.76              & 126              & 70.63                                                             \\ \hline
\textbf{Manipuri}                                               & 0.62               & 0.25            & 0.35              & 53               & 24.52                                                             \\ \hline
\textbf{Mohiniattam}                                            & 0.97               & 0.94            & 0.95              & 63               & 93.63                                                             \\ \hline
\textbf{Odissi}                                                 & 0.83               & 0.64            & 0.72              & 69               & 63.76                                                             \\ \hline
\textbf{Average}                                                & 0.75               & 0.72            & 0.71              & 445              & \textbf{72.35}                                                    \\ \hline
\end{tabular}
\end{table}

\begin{table*}[]
\label{tab:2}
\caption{Comparison of various features and their combinations for Dance Classification}
\begin{tabular}{|c|c|c|c|c|c|c|c|}
\hline
\textbf{Method}                                                                            & \textbf{Bharatnatyam} & \textbf{Kathak} & \textbf{Kuchipudi} & \textbf{Manipuri} & \textbf{Mohiniattam} & \textbf{Odissi} & \textbf{\begin{tabular}[c]{@{}c@{}}Average \\ Accuracy\end{tabular}} \\ \hline
\textbf{InceptionV3}                                                                       & 80.48                 & 62.71           & 28.90              & 30.64             & 98.41                & 53.62           & 59.1                                                                 \\ \hline
\textbf{Pose Signature}                                                                    & 88.15                 & 79.31           & 43.65              & 41.50             & 96.82                & 71.01           & 67.41                                                                \\ \hline
\textbf{Kinetics}                                                                          & 84.21                 & 68.96           & 56.34              & 30.18             & 96.82                & 57.97           & 65.61                                                                \\ \hline
\textbf{\begin{tabular}[c]{@{}c@{}}InceptionV3+\\ Pose Signature\end{tabular}}             & 51.31                 & 67.24           & 65.87              & 67.92             & 93.65                & 73.91           & 68.98                                                                \\ \hline
\textbf{\begin{tabular}[c]{@{}c@{}}InceptionV3+\\ Kinetics\end{tabular}}                   & 85.52                 & 68.96           & 63.49              & 24.53             & 96.82                & 57.97           & 67.19                                                                \\ \hline
\textbf{\begin{tabular}[c]{@{}c@{}}InceptionV3+\\ Pose Signature+\\ Kinetics\end{tabular}} & 86.84                 & 87.93           & 70.63              & 24.52             & 93.65                & 63.76           & 72.35                                                                \\ \hline
\end{tabular}
\end{table*}
We have also shown the evaluation on various possible permutations of the feature sets to validate the efficacy of the concatenation on improving discriminative ability without hindering the network's training performance. Tab. 4 provides an evaluation on various possible combination of the feature sets. In order to further improve the feature representation and model the spatio-temporal dependencies, we incorporate Residual Network with multiple cardinalities (also known as ResNext) using 3D CNN architecture. The ResNext-101 was trained on Kinetics dataset, which comprises of $400$ video classes. The ResNext architecture utilizes 16 consecutive frames as input and outputs 3D feature vector with size $2048$ (or $16$x$128$). We implemented ResNext for all frames in the videos (with sequence length>$48$). The output is then saved in a JSON format with segment length $16$ per video. We convert this data into a feature vector of $128$ dimensionality for the input images and then selected $48$ frames again uniformly for training. The training input is of size ($32$,$48$,$128$). The training is computationally less expensive due to reduced feature vector. The results were better than InceptionV3 which utilised only spatial information, but since ResNext captures flow i.e. temporal information, the results were better even with reduced feature length.

Next, we utilize AlphaPose framework to compute pose of the dancing person. The pose constitutes $16$ anchor joints of different body parts. We then utilize these $16$ anchor joints to construct a pose signature capturing the temporal flow and flow directionality between successive frames and embed the geometric constraints with the normalized distances between various body anchor joints. The resultant pose signature is of dimensionality $75-D$ feature vector. LSTM achieved the best results which is comparable to the result using ResNext-101 3D CNN feature used independently. Since both the architectures reflect temporal information, the results differ by close margins.

After training these features individually, we concatenate these features to construct a novel feature set for training via LSTM networks. The InceptionV3 combined with kinetics gave slightly better results ($\sim1-2\%$) than combined with pose signature. The combined feature descriptor ($2251$ length) gives the best performance.

\section{Conclusion}
\label{sec:conclusion}
We have presented a novel pose signature in a sequencial learning framework for Indian Classical Dance (ICD) classification. We incorporated pose, flow and spatio-temporal dependencies in the pose signature to capture the adjacency relationship between anchor joints in the skeletal pattern of the dancer. We performed exhaustive experiments and demonstrated the effectiveness of the proposed methodology on dance classification. 
We showed that deep descriptors with handcrafted pose signature outperformed on ICD dataset. We also showed that due to high similarities between dance moves and dressing attires it is highly challenging to classify dance sequences. In future works, we plan to incorporate facial gestures into the classification pipeline.

%
\bibliographystyle{abbrv}
\bibliography{sigproc}  
%
%


\end{document}